\documentclass[conference]{IEEEtran}
\IEEEoverridecommandlockouts
\usepackage{hyperref}
\usepackage{cite}
\usepackage{amsmath,amssymb,amsfonts}
\usepackage{graphicx}
\usepackage{textcomp}
\usepackage{xcolor}
\def\BibTeX{{\rm B\kern-.05em{\sc i\kern-.025em b}\kern-.08em
    T\kern-.1667em\lower.7ex\hbox{E}\kern-.125emX}}

\usepackage{algorithm}
\usepackage{algpseudocode}

\newcommand{\C}{\mathbb C}

\newcommand{\Dd}{\mathbf D}

\newcommand{\sd}{\mathbf{s}}                        

\newcommand{\XX}{\mathbf{x}}

\newcommand{\ZZ}{\mathbf{z}}

\newcommand{\YY}{\mathbf{y}}

\newcommand{\Ad}{\mathbf A}                        
 
\newcommand{\Fd}{\mathbf F} 
\newcommand{\Bd}{\mathbf B}

\newcommand{\Su}{\mathbf{S}_I}

\newcommand{\herm}{{\scriptstyle \boldsymbol{\mathsf{H}}}}
\newcommand{\trans}{{\scriptstyle \boldsymbol{\mathsf{T}}}}

\newcommand{\LLambda}{\boldsymbol{\Lambda}}

\begin{document}

\title{Learning Spatially Adaptive $\ell_1$-Norms Weights for Convolutional Synthesis Regularization\\
\thanks{The project (22HLT02 A4IM) has received funding from the European Partnership on Metrology, co-financed from the European Union’s Horizon Europe Research and Innovation Programme and by the Participating States. LC acknowledges the financial support of the European Research Council (grant MALIN, 101117133).}
}

\makeatletter
\newcommand{\linebreakand}{%
  \end{@IEEEauthorhalign}
  \hfill\mbox{}\par
  \mbox{}\hfill\begin{@IEEEauthorhalign}
}
\makeatother

\author{\IEEEauthorblockN{
Andreas Kofler}
\IEEEauthorblockA{
\textit{Physikalisch-Technische Bundesanstalt (PTB)}\\
 Braunschweig and Berlin, Germany \\
andreas.kofler@ptb.de}
\and
\IEEEauthorblockN{
Luca Calatroni}
\IEEEauthorblockA{\textit{MaLGa Center, DIBRIS,} \\
\textit{Università di Genova},\\
 \textit{MMS, Istituto Italiano di Tecnologia}, \\
Genoa, Italy \\
luca.calatroni@unige.it \vspace{1em}}
 \linebreakand 
\IEEEauthorblockN{
Christoph Kolbitsch}
\IEEEauthorblockA{
\textit{Physikalisch-Technische Bundesanstalt (PTB)}\\
Braunschweig and Berlin, Germany \\
christoph.kolbitsch@ptb.de}
\and
\IEEEauthorblockN{
Kostas Papafitsoros}
\IEEEauthorblockA{\textit{School of Mathematical Sciences} \\
\textit{Queen Mary University of London}\\
London, UK \\
k.papafitsoros@qmul.ac.uk}
}

\maketitle

\begin{abstract}
 We propose an unrolled algorithm approach for learning spatially adaptive parameter maps in the framework of convolutional synthesis-based $\ell_1$ regularization. More precisely, we consider a family of pre-trained convolutional filters and estimate deeply parametrized spatially varying parameters applied to the sparse feature maps by means of unrolling a FISTA algorithm to solve the underlying sparse estimation problem. The proposed approach is evaluated for image reconstruction of low-field MRI and compared to spatially adaptive and non-adaptive analysis-type procedures relying on Total Variation regularization and to a well-established model-based deep learning approach. We show that the proposed approach produces visually and quantitatively comparable results with the latter approaches and at the same time remains highly interpretable. In particular, the inferred parameter maps quantify  
 the local contribution of each filter in the reconstruction, which provides valuable insight into the algorithm mechanism and could potentially be used to discard unsuited filters.
\end{abstract}


\section{Introduction}
Convolutional Sparse Coding (CSC) and Convolutional Dictionary Learning (CDL) approaches  \cite{garcia2018convolutional} rely on the assumption that the signal/image of interest $\XX\in\mathbb{R}^N$ (or $\mathbb{C}^N$) is well-approximated by a linear combination of convolutions of normalized filters $d_k \in \mathbb{R}^{k_f\times k_f}$ with sparse feature maps $\sd_k \in \mathbb{R}^{N}$ (or $\mathbb{C}^{N}$) for $k=1,\ldots, K$, that is
\begin{equation}\label{eq:cdl_model}
    \XX \approx \sum_{k=1}^K d_k \ast \sd_k, \quad \text{ where } \quad \forall k \quad \sd_k \,\text{ is sparse}.
\end{equation}
By enforcing sparsity by means of $\ell_1$-regularization, given a set of signals $\{\XX_l\}_{l=1}^L$ and some $\lambda>0$, the CDL problem can be thus typically formulated as:
\begin{equation}\label{eq:cdl_csc}
\begin{aligned}
    \underset{\{d_k\}, \{\sd_k\}}{\min} \; &\frac{1}{2}\sum_{l=1}^L \|\XX_l - \sum_{k=1}^K d_k \ast \sd_{k,l}\|_2^2 + 
    \lambda{\sum_{l=1}^L\sum_{k=1}^K}\|\sd_{k,l} \|_1 \\  &\text{such  that } \quad  \|d_k\|_2 = 1, \quad \forall k = 1, \dots, K.
\end{aligned}
\end{equation}
In the context of image reconstruction, given possibly incomplete observed data $\YY\in V$ modeled as the noisy output of a linear forward operator $\Ad:V \rightarrow W$, CDL and CSC are employed as variational regularization methods, see e.g.\ \cite{quan2016compressed}:
\begin{equation} \label{eq:image_recon_with_csc_cdl}
    \begin{aligned}
        \underset{\XX, \{d_k\}, \{\sd_k\}}{\min} \; & 
        \frac{1}{2} \|\Ad\XX - \YY\|_2^2 +\\
         \frac{\alpha}{2} & \|\XX - \sum_{k=1}^K d_k \ast \sd_k\|_2^2 
        +  \lambda\sum_{k=1}^K \|\sd_k\|_1 \\
        \text{such  that } \quad & \|d_k\|_2 = 1, \quad \forall k = 1, \dots, K,
    \end{aligned}
\end{equation}
where $\alpha>0$ enforces the synthesis constraint.
Note that in \eqref{eq:image_recon_with_csc_cdl} it is implicitly assumed that the underlying sparsifying model is unknown so that the elements $\XX, \{d_k\}, \{\sd_k\}$, $k=1\ldots,K$ are reconstructed jointly, typically by alternating the minimization between solving \eqref{eq:cdl_csc} and computing the current update of desired image depending on the estimated dictionary.
These methods are thus often referred to as \textit{blind} Compressed Sensing approaches, see, e.g.~\cite{Lingala2013} for MRI applications.

Problem \eqref{eq:image_recon_with_csc_cdl} is a non-convex optimization problem whose solution is, typically,  computationally demanding, and further requires careful tuning of the regularization parameters $\lambda$, $\alpha$, initializations $\XX^0, \{d^0_k\}, \{\sd^0_k\}$ and, depending on the algorithms considered, possibly also of other parameters coupling the image update and the CSC and CDL, see, e.g.\ \cite{wohlberg2015efficient} for ADMM schemes. Furthermore, the regularization parameter $\lambda$ dictates the strength of the imposed regularization in terms of the sparsity of the feature maps only globally and independently on the specific filter considered. This can be limiting, as the filter-dependent representation of local image details and structures may vary, asking for adaptive weighting of the feature maps, similarly as in\ \cite{pali2021adaptive} where this issue is addressed within a non-convolutional setting (i.e.\ patch-based dictionary learning and sparse coding).\\

\vspace{-0.3cm}

\textbf{Contribution.} 
In this work, we propose a synthesis-based spatially adaptive reguralization scheme. We structure our proposal around two main modules.  First, we learn adaptive parameter maps that weight the convolutional sparse approximation using pre-learned filters. Then, we solve the corresponding weighted-$\ell_1$ problem using the fast iterative shrinkage algorithm (FISTA) \cite{beck2009fast}. By utilizing pre-computed filters, our formulation avoids the need for alternating minimization while remaining sample-adaptive, as the regularization strength is estimated at inference for each specific sample. 
Our work is inspired by a recently proposed approach \cite{kofler2023learning} which uses adaptive spatio-temporal Total Variation (TV) regularization maps parametrized by convolutional neural networks (CNNs), which are learned in a supervised fashion through unrolling  
of the primal-dual hybrid-gradient (PDHG) algorithm \cite{monga2021algorithm, chambolle2011first} ; see also \cite{vu2025deep_published} for a recent extension to Total Generalized Variation.
Note that employing standard finite-difference discretization filters of the gradient operator corresponds to enforcing a sparsity-based approach with fixed hand-crafted filters within an analysis setting. 

We report several numerical results for exemplar low-field MRI data, demonstrating quantitative improvements over non-adaptive models and providing quantitative indications (encoded in the estimated maps) about the adequacy of individual filters for signal representation.

\section{Methods} 
We provide details on the chosen problem formulation, the reconstruction algorithm used and the neural networks used to estimate the parameter maps. 

\subsection{Problem Formulation}

We assume to have access to a set of pre-trained unit-norm convolutional dictionary filters $\{d_k\}_{k=1}^K$ with $d_k\in\mathbb{R}^{k_f\times k_f}$ for all $k$. Instead of relying on the use of a quadratic term weighted by $\alpha$ to enforce \eqref{eq:cdl_model}  as in \eqref{eq:image_recon_with_csc_cdl}, we explicitly enforce \eqref{eq:cdl_model} in terms of a re-parametrization of the underlying reconstruction problem. To make the notation more compact, we thus define:
\begin{eqnarray}\label{eq:notation}
    \sd := [\sd_1, \ldots, \sd_K]^\trans,\quad 
    \Dd \sd := \sum_{k=1}^K d_k \ast \sd_k,\quad 
    \Bd := \Ad \Dd,
\end{eqnarray}
and then consider the problem
\begin{equation}\label{eq:cdl_explicit_lambda_scalar}
    \underset{\sd \in \mathbb{R}^{NK} }{\min}~~ \frac{1}{2}\|\Bd \sd - \YY\|_2^2 + \lambda \| \sd\|_1,
\end{equation}
where $\lambda$ \textit{globally} dictates the sparsity level of \textit{all} $K$ feature maps $\{\sd_k\}_k$.
To adapt the regularization strength both locally as well as differently for each $\sd_k$, we now consider the adaptive version given by:
\begin{equation}\label{eq:cdl_explicit_lambda_map}
    \underset{\sd}{\min}\,  \frac{1}{2}\|\Bd \sd - \YY\|_2^2 +  \| \boldsymbol{\LLambda} \sd\|_1,
\end{equation}
with $\LLambda\in\mathbb{R}^{NK \times NK}$ being a block-diagonal operator with strictly positive entries, i.e.\ $\LLambda:= \mathrm{diag}(\LLambda_1, \ldots, \LLambda_K)$, so that the term $\| \boldsymbol{\LLambda} \sd\|_1$ can be written as 
$\| \boldsymbol{\LLambda} \sd\|_1:=\sum_{k=1}^K\| \sd_k \|_{\LLambda_k,1}:= \sum_{k=1}^K\sum_{j=1}^{N} \big|\LLambda_k[j] \cdot \sd_k[j] \big|_1$. To compute such adaptive $\ell_1$ maps, we will make use of an algorithm unrolling strategy. Note that in \cite{Pourya2025} similar ideas for adaptive least-square regularizers are used.

\subsection{Algorithm Unrolling}
To learn the parameter maps $\{\LLambda_k\}_{k=1}^K$, we unroll $T$ iterations of the well-known accelerated proximal gradient descent method FISTA \cite{beck2009fast}. In particular, we use the version of FISTA presented in \cite{chambolle2015convergence}, which ensures the convergence of the iterates to a minimizer of problem \eqref{eq:cdl_explicit_lambda_map}.
Note that in the FISTA update, the proximal operator of $\tau g(\sd):=\tau \| \boldsymbol{\LLambda} \sd\|_1$ with $\tau>0$ denoting the algorithmic step-size, can be explicitly computed by $\mathrm{prox}_g(\ZZ)=\mathcal{S}_{\tau \LLambda}(\ZZ)$, the componentwise soft-thresholding operator with thresholds defined by the respective values of the corresponding $\LLambda$-map.
Furthermore, convergence is guaranteed under the choice $\tau=1/ \|\Bd\|^2$. As an alternative, iterative rewweighted-$\ell_1$ strategies can be used as in \cite{CandesIRL1}, with the disadvantage of being less interpretable from the underlying variational perspective.

\begin{figure*}[h]
    \centering
    \includegraphics[width=0.98\linewidth]{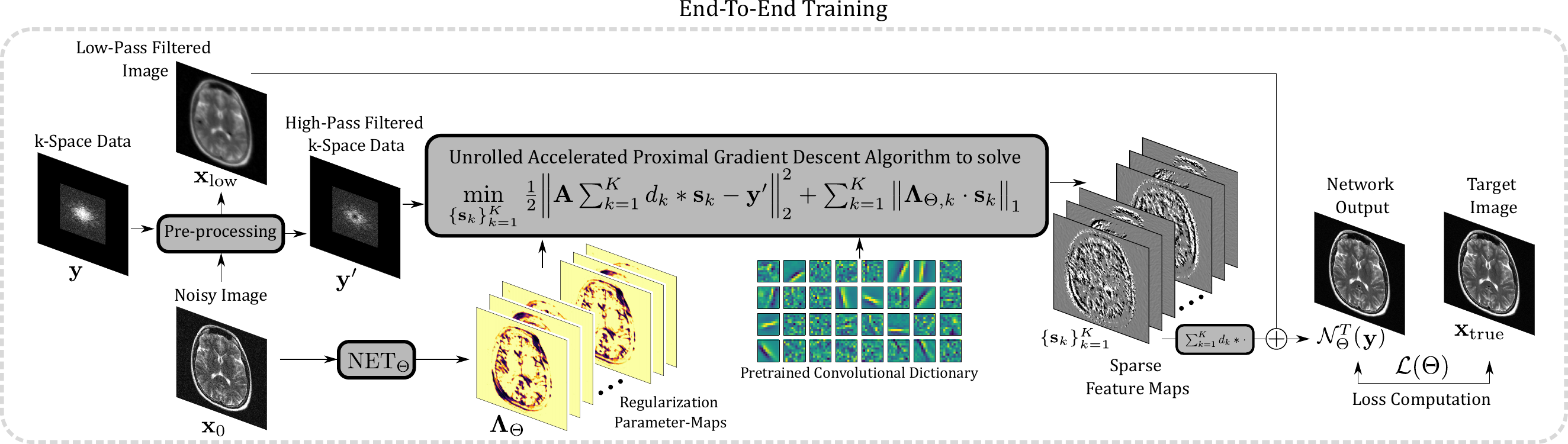}
    \caption{The proposed pipeline consists of three blocks: the first one is  a high-pass pre-processing to
construct the raw data $\YY^\prime$. The second defines a re-parametrization of adaptive $\{\LLambda_k\}_k$-maps in terms of a CNN $\text{NET}_\Theta$ given the input image $\XX_0:=\Ad^\herm \YY$. The third consists of unrolling $T$ iterations of the FISTA algorithm for solving \eqref{eq:cdl_explicit_lambda_map_high_passed}. Note that the pre-processing block includes a parameter $\beta>0$, which is learned together with the network weights $\Theta$.}
    \label{fig:cdl_approach}
\end{figure*}

\subsection{High-Pass Filtering} \label{sec:HP_filter}
Most uses of sparse coding approaches require the application of high-pass filtering for better performance, see e.g.\ \cite{wohlberg2016convolutional}. We thus consider a pre-processing of the data aiming to learn the $\{\LLambda_k\}_k$-maps for the sparse representation of a high-passed filtered representation of the sought image. For that, we employ simple filtering based on gradient regularization as described in \cite{wohlberg2016convolutional}. Namely, for an initial image estimate $\XX_0$, we define the $\beta$-high-passed image $\XX_{\mathrm{high}}$ by 
$  \XX_{\mathrm{high}}:= \XX_0 - \XX_{\mathrm{low}}$
where the low-pass component $\XX_{\mathrm{low}}$ is obtained by
\begin{equation}\label{eq:hihg_pass_filtering}
     \XX_{\mathrm{low}}:= 
     \underset{\XX}{\mathrm{argmin}}\;
      \frac{1}{2}\| \XX - \XX_{0}\|_2^2 + \frac{\beta}{2}\| \nabla \XX\|_2^2,
\end{equation}
via solving a linear system, by means, e.g., of the conjugate gradient (CG) method. 
Then, we enforce \eqref{eq:cdl_model} on the image $\XX_{\mathrm{high}}$, define $\YY^\prime:=\YY - \Ad \XX_{\mathrm{low}}$  and consider
\begin{equation}\label{eq:cdl_explicit_lambda_map_high_passed}
    \underset{\sd}{\min}\,  \frac{1}{2}\|\Bd\sd - \YY^\prime\|_2^2 +  \| \boldsymbol{\LLambda} \sd\|_1,
\end{equation}
where thus 
the convolutional dictionary $\Dd$ is required to approximate the high-passed component only, i.e.\ $\Dd \sd\approx\XX_{\mathrm{high}}$ and the term $\Ad \XX_{\mathrm{low}}$ is absorbed in $\YY^\prime$.\\
Further, in contrast to \cite{kofler2023learning},  we enforce the parameter maps $\{\LLambda_k\}_k$ to have a  component-wise upper bound to avoid the scaling ambiguity between the optimization variables $\sd$ and the parameter maps $\{\LLambda_k\}_k$.

\subsection{Network Architecture}

To obtain the $\{\LLambda_k\}_k$-maps, we follow \cite{kofler2023learning}. 
Our network consists of three blocks: $\mathrm{i})$ a high-pass pre-processing to construct the raw data $\YY^\prime$ in \eqref{eq:cdl_explicit_lambda_map_high_passed}, $\mathrm{ii})$ a deep-network module which outputs the $K$ parameter maps $\{\LLambda_k\}_k$ from a given initial image and $\mathrm{iii})$ the unrolling of $T$ iterations of FISTA to approximately solve problem \eqref{eq:cdl_explicit_lambda_map_high_passed}.

The parameter maps $\{\LLambda_k\}_k$ are re-parametrized as the output of an expressive CNN with $K$ output-channels which takes as input an initial image estimate, e.g.\ $\XX_0:=\Ad^\herm \YY$ or $\XX_0:=\Ad^\dagger \YY$. 
In particular, upon vectorization of  2D images into vectors for $k=1,\ldots,K$, we have that: 
\begin{equation}\label{eq:NET}
    \LLambda_k:=(\mathrm{NET}_{\Theta}(\XX_0))_k:= t\cdot\boldsymbol{\phi}\, \circ \left(\mathbf{u}_{\Theta}
    (\XX_0)\right)_k,
\end{equation}
where $\mathbf{u}_{\Theta}$ denotes a 2D U-Net \cite{ronneberger2015u} with parameters $\Theta$ and $K$ (vectorized) output channels, $\boldsymbol{\phi}$ is the sigmoid activation function acting on all output channels component-wise, and $t>0$ is a hyper-parameter controlling the upper bound of the obtainable $\{\LLambda_k\}_k$-maps. \\
For constructing $\YY^\prime$ to be used in the FISTA-block as described in \ref{sec:HP_filter}, a CG-module solving \eqref{eq:hihg_pass_filtering} is applied. The parameter $\beta$ in \eqref{eq:hihg_pass_filtering} can thus be learned as well.\\
The third block unrolls $T$ iterations of FISTA to estimate the solution $\sd^{\ast}$ of problem \eqref{eq:cdl_explicit_lambda_map_high_passed}, from which then, an estimate of the image can be retrieved as $\XX^\ast:= \Dd\sd^{\ast} + \XX_{\mathrm{low}}$.
Figure \ref{fig:cdl_approach} illustrates the entire pipeline, which we denote as $\mathcal{N}^T_{\Theta}$.\\

\section{Experiments}

\subsection{Dataset} 
 We apply the proposed method to 2D low-field brain MR imaging. Low-field MRI suffers from high noise and low image resolution due to hardware constraints. In contrast to standard MRI, only a single receiver coil is used for data acquisition. Using the notation above, we set here $V=W=\C^N$, $\Ad:=\Su \Fd$,  where $\Fd:\C^N \rightarrow \C^N$ denotes the 2D Fourier-transform and $\Su:\C^N \rightarrow\C^N$ denotes a binary mask that only retains $k$-space coefficients indexed by $I\subset J=\{1,\ldots, N\}$ and masks out measurements corresponding to high frequencies leading to low-resolution images. The measurements $\YY\in\C^N$ are corrupted by additive white Gaussian noise realizations $\mathbf{e}\in \C^N$ with standard deviation $\sigma$. This problem is ill-posed due to the sub-sampling mask $\Su$ and the low SNR given by the presence of noise which, in practice, is typically relatively high. For realistic experiments, we thus chose   $\sigma \in \{0.075, 0.15, 0.3\}$.\\
 We rely on supervised training to train the proposed network. For training, validation, and testing, we used a total of 6917 brain MR images taken from the fastMRI dataset \cite{zbontar2018fastmri}. We split these into 4875/1393/649 images for training, validation, and testing respectively. To obtain complex-valued images, coil-sensitivity maps were estimated from the multi-coil $k$-space data, from which a coil-combined image was reconstructed. For this step, we used the method in \cite{walsh2000adaptive} implemented in \href{https://github.com/PTB-MR/mrpro}{\texttt{MRpro}} \cite{zimmermann2025}.

 \subsection{Comparisons}
To compare our results, we considered the well-known analysis-type TV-regularized problem with scalar regularization parameter (TV-$\lambda$) as well as its weighted extension where spatially varying regularization parameter maps are learned as in \cite{kofler2023learning} (TV-$\LLambda$). As a further comparison, we considered the model-based deep learning (MoDL) approach \cite{aggarwal2018modl}, which can be derived by learning the proximal operator of the regularizer by means of a variable splitting method and a deep CNN parametrization. 
Note that our approach is more interpretable than  MoDL, since, once the filters $\{d_k\}_k$ and the $\{\LLambda_k\}_k$-maps are learned, the reconstruction computed is  a minimizer to the functional \eqref{eq:cdl_explicit_lambda_map_high_passed}. The use of the variant of FISTA in \cite{chambolle2015convergence} guarantees that convergence is guaranteed at a numerical level too. Analogously to TV-$\LLambda$, our CDL approach is referred to as CDL-$\LLambda$, while CDL-$\lambda$ refers to the approach used to solve problem \eqref{eq:cdl_explicit_lambda_scalar} with one global (non-adaptive) scalar $\lambda>0$.
We evaluate the results in terms of the structural similarity index measure (SSIM) and peak signal-to-noise ratio (PSNR), which are computed after constraining the images on regions excluding the background where no MR signal is present by means of adequately defined binary masks derived from the target images.

\begin{figure*}[h]
    \centering
    \includegraphics[width=0.9\linewidth]{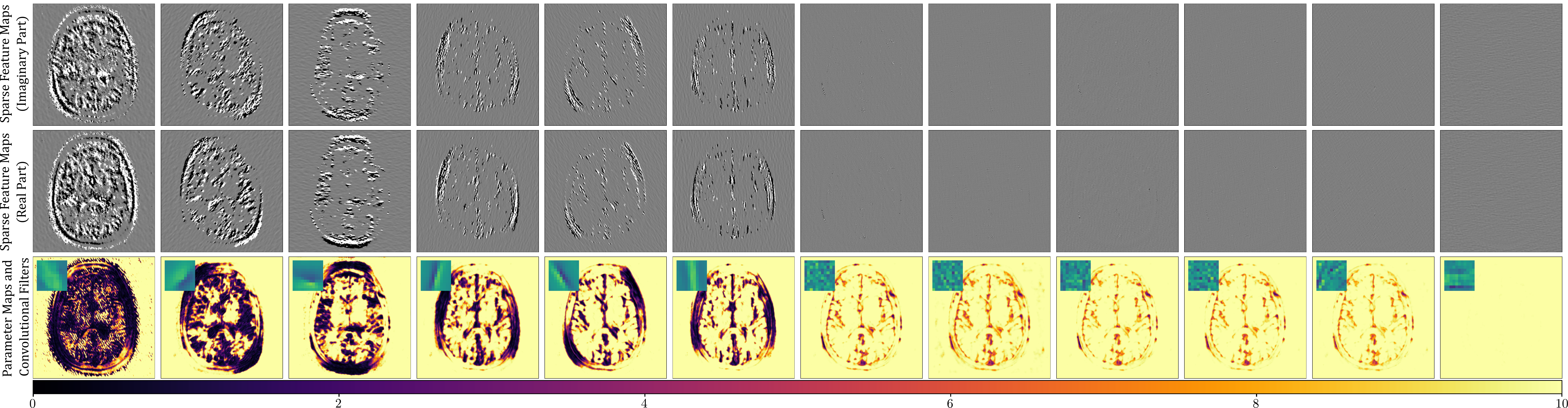}
    \caption{An example of 12 out of $K=64$ $\{\LLambda_k\}_k$-maps for dictionary filters with filter size $k_f \times k_f = 11\times 11$ along with the corresponding sparse feature maps. 
    The first two rows show the feature maps of the real and imaginary part of the solution of \eqref{eq:cdl_explicit_lambda_map_high_passed}, while the third row shows the corresponding $\{\LLambda_k\}_k$-maps.
    Results are ordered in a decreasing order w.r.t.~the variance of the $\{\LLambda_k\}_k$-maps. 
    The results indicate that the network $\mathrm{NET}_{\Theta}$ can provide structural information associated with each filter and, moreover, can identify filters that are less relevant for the reconstruction (high values of the map).}
    \label{fig:sparse_codes_maps_n_filters}
\end{figure*}

\begin{figure*}[h]
    \centering
    \includegraphics[width=\linewidth]{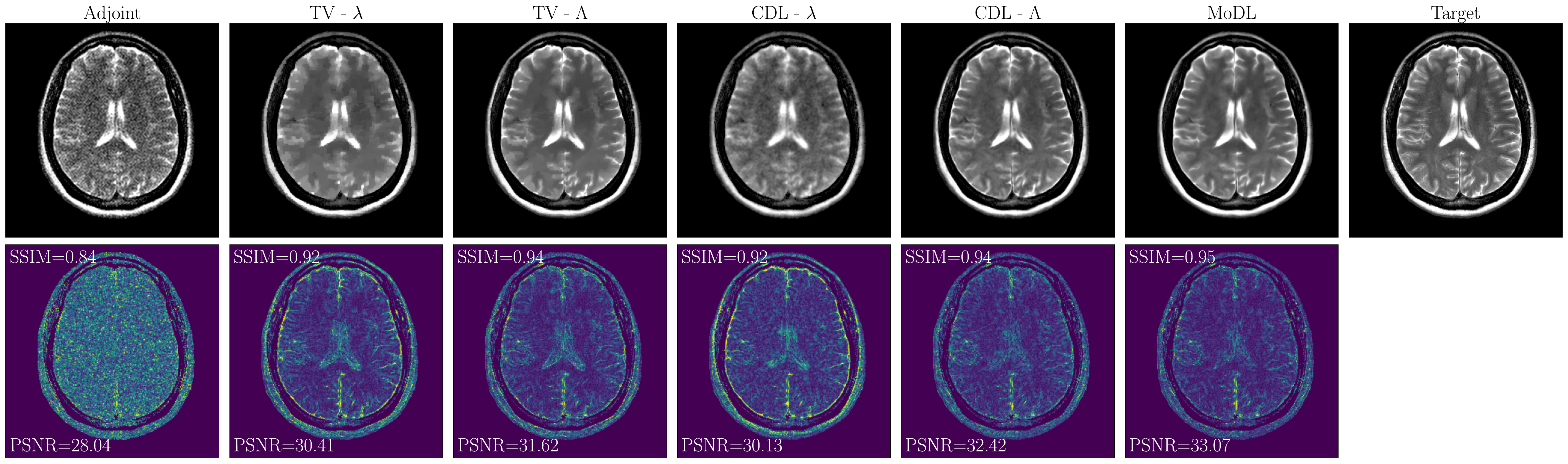}
    \caption{An example of images reconstructed with the different methods for a noise level of $\sigma=0.3$. The SSIM and the PSNR, which are computed by masking out values outside of the defined mask, where no signal is present, are listed in the point-wise error images. To better highlight residual errors that can be attributed to the model limitations, the point-wise error images were scaled by a factor of 3.
    }
    \label{fig:recon_results}
\end{figure*}

\subsection{Training and Implementation Details}
The 2D U-Net $\mathbf{u}_{\Theta}$ used in the CNN-block $\mathrm{NET}_{\Theta}$ in \eqref{eq:NET} has a 2-$K$-$2K$-$4K$-$2K$-$K$ structure with two input-channels (for the real and imaginary parts of the image) and $K$ output channels for $K$ parameter maps $\left\{\mathbf{\Lambda}_k\right\}_k$. To handle complex-valued data, note that the real and the imaginary parts of the images are approximated using the same filters, i.e.\ we have $\Dd \sd:= \sum_k (d_k \ast \mathrm{Re}(\sd_k) + \mathrm{i} \,d_k\ast \mathrm{Im}(\sd_k))$.
For training, we used the Adam optimizer \cite{kingma2014adam} with an MSE-loss using a batch-size of 2. For all methods, the scalar regularization parameters ($\lambda$ and $\beta$), which were "activated" using a Softplus activation function to maintain strict positivity, were assigned a learning rate of $10^{-1}$, while the U-Nets $\mathbf{u}_{\Theta}$ were given a learning rate of $10^{-4}$ with a weight decay regularization weight of $10^{-5}$.\\
The dictionary filters $\{d_k\}_k$ were pre-trained on 160 randomly selected images from the training set using the algorithm described in \cite{liu2018first} implemented in SPORCO \cite{wohlberg2017sporco}. We tested different kernel sizes $k_f\times k_f = 7 \times 7$, $11\times 11$ and $13\times 13$ as well as different numbers of filters $K=32,64$. Due to space limitations, we here only report results for $K=64$ and $k_f\times k_f= 11 \times 11$. We point out that the experiments suggest that more and larger filters tend to improve the achievable performance. The upper bound $t>0$ in \eqref{eq:NET} was fixed to $t=10.0$, which is also the average maximal magnitude value the images were normalized to.\\
The number of unrolling steps was set to $T=128$ for TV-$\lambda$ and TV-$\LLambda$,  to $T=64$ for CDL-$\lambda$ and CDL-$\LLambda$ (which enjoy faster convergence inherited from FISTA), and to $T=8$ for MoDL. The number of epochs was 96 for MoDL ($\approx 16$\,h training), 4 for TV-$\lambda$ and CDL-$\lambda$ ($\approx 4$\,h training time) 48 for TV-$\LLambda$ ($\approx 11$\,h training time) and CDL-$\LLambda$ ($\approx  42$\,h training time).
Upon integration of all necessary building blocks in \href{https://github.com/PTB-MR/mrpro}{\texttt{MRpro}} \cite{zimmermann2025}, the code of this work will be made available at \href{https://github.com/koflera/LearningL1NormsWeights4Synthesis}{github.com/koflera/LearningL1NormsWeights4Synthesis}.

\section{Results and Discussion}

In Figure \ref{fig:sparse_codes_maps_n_filters} we see an example of 12 sparse feature maps estimated by FISTA as well as the corresponding learned regularization parameter maps and their associated filters. From the parameter maps, we see that the network seems to be able to distinguish between more and less useful filters for the sparse approximation. Note that higher values in the $\{\LLambda_k\}_k$-maps result in a higher threshold in the proximal operator of $g$ and thus to a lower contribution in the sparse representation. 

\begin{figure}[t]
    \centering
    \includegraphics[width=0.8\linewidth]{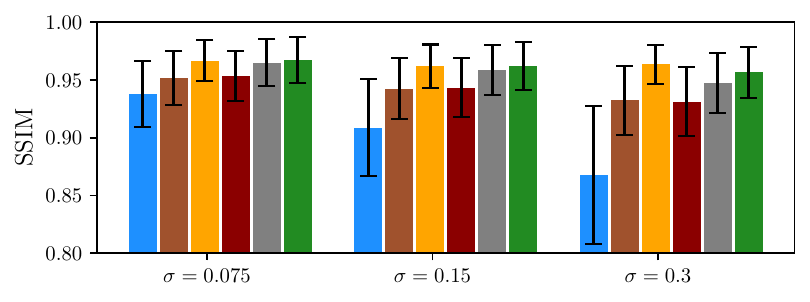}
    \includegraphics[width=0.8\linewidth]{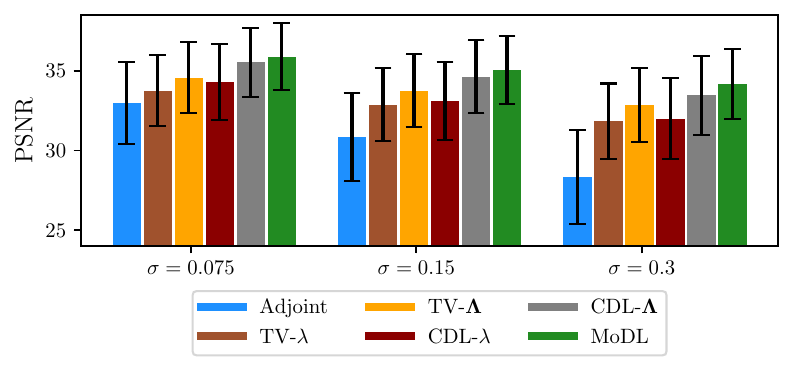}
    \caption{A summary of SSIM and PSNR obtained on the test set by the different approaches. The proposed CDL-$\LLambda$ achieves competitive results compared to the TV-$\LLambda$ \cite{kofler2023learning} as well as to the deep learning-based method MoDL \cite{aggarwal2018modl}.}
    \label{fig:box_plots}
\end{figure}

Figure \ref{fig:recon_results} shows an example of images reconstructed for a noise level of $\sigma=0.3$. All methods successfully improved upon the adjoint-based reconstruction, which heavily suffers from noise and blurring artifacts.
The methods employing only scalar regularization values $\lambda$ exhibit larger errors at the boundaries and are clearly outperformed by both TV-$\LLambda$ \cite{kofler2023learning} and the proposed CDL-$\LLambda$, which employ learned regularization parameter maps and yield results comparable to the deep learning-based method MoDL \cite{aggarwal2018modl}. 
Although similar in terms of SSIM and PSNR, the investigated CDL-$\LLambda$  yields more anatomically accurate images in contrast to TV-$\LLambda$, which, despite the spatially adaptive regularization, tends to exhibit some piece-wise constant structures. Finally, CDL-$\LLambda$ exhibits a high visual similarity when compared to MoDL.

Figure \ref{fig:box_plots} summarizes the results obtained on the test set in terms of SSIM and PSNR. The methods involving only scalar regularization parameters (TV-$\lambda$ and CDL-$\lambda$) are outperformed by both TV-$\LLambda$ and CDL-$\LLambda$ involving the learned adaptive spatial regularization, which are competitive even when compared to the deep learning-based MoDL \cite{aggarwal2018modl}.

\section{Conclusion}

In this work, we extended the method proposed in \cite{kofler2023learning} for learning TV adaptive maps via CNN parametrization and algorithmic unrolling to the sparse convolutional synthesis setting. The proposed approach combines model-based regularization, in our case CSC and CDL, with deep neural networks;
it yields competitive results compared to MoDL \cite{aggarwal2018modl}, but further enjoys interpretability, convergence guarantees, and informative quantitative estimation of the local behavior with respect to a pre-trained set of convolutional filters.
As a byproduct, the learned parameter maps can be used to discriminate between more and less useful convolutional filters in the representation of the solution, resulting in an at-inference-time self-adaptation of the reconstruction method. As a consequence, this approach offers a valuable alternative to the well-established paradigm of learning new filters jointly with the desired solution for each reconstruction problem at hand.\\
A possible future direction of this work could be the use of self-supervised training procedures to estimate adequate regularization maps without the need for reference data.

\bibliographystyle{IEEEtran}
\bibliography{biblio}

\end{document}